\documentclass[10pt,twocolumn,letterpaper]{article}

\usepackage{3dv}
\usepackage{times}
\usepackage{epsfig}
\usepackage{graphicx}
\usepackage{amsmath}
\usepackage{amssymb}
\usepackage[usenames]{xcolor}
\usepackage{ifthen}
\usepackage{float}

\definecolor{brilliantrose}{rgb}{1.0, 0.33, 0.64}
\definecolor{amber}{rgb}{1.0, 0.49, 0.0}
\newcommand{\PG}[1]{{{#1}}} 

\newboolean{showComments}
\setboolean{showComments}{true}
\definecolor{gold}{rgb}{0.80,.60,0}

\newcommand{\todo}[2]{
\ifthenelse{\boolean{showComments}}
{\framebox[\columnwidth][l]{\parbox[l]{3.25in}{\textcolor{red}{#1: #2}}}}
{}
}

\newcommand{\norm}[1]{\left\lVert#1\right\rVert}

\newcommand{\modelnamenbsp}{Dropout Autoencoder}
\newcommand{\modelname}{\modelnamenbsp~}

\newcommand{\modelnamefullnbsp}{\modelname LSTM}
\newcommand{\modelnamefull}{\modelnamefullnbsp~}

\newcommand{\DAE}{DAE}
\newcommand{\LSTMpart}{LSTM3LR}
\newcommand{\DAELSTMthreeLR}{DAE-LSTM}

\newcommand{\DAEspace}{\DAE~}
\newcommand{\LSTMpartSpace}{\LSTMpart~}

\usepackage[pagebackref=true,breaklinks=true,letterpaper=true,colorlinks,bookmarks=false]{hyperref}

\threedvfinalcopy 
\newif\ifmainpaper
\mainpapertrue 

\def\threedvPaperID{95} 

\ifthreedvfinal\fi
\begin{document}

\ifmainpaper
\title{Learning Human Motion Models for Long-term Predictions}

\author{Partha Ghosh\quad \quad Jie Song \quad \quad Emre Aksan \quad \quad Otmar Hilliges\\
Advanced Interactive Technologies, ETH Zurich\\
{\tt\small pghosh@student.ethz.ch \quad \{jsong, eaksan, otmar.hilliges\}@inf.ethz.ch \quad }
}

\maketitle

\begin{abstract}
   We propose a new architecture for the learning of predictive spatio-temporal motion models from data alone. Our approach, dubbed the \modelnamefull (\DAELSTMthreeLR), is capable of synthesizing natural looking motion sequences over long-time horizons\footnote{$>10s$ for periodic motions, e.g. walking, $>2s$ for aperiodic motion, e.g. eating} without catastrophic drift or motion degradation. The model consists of two components, a 3-layer recurrent neural network to model temporal aspects and a novel autoencoder that is trained to implicitly recover the spatial structure of the human skeleton via randomly removing information about joints during training. This \modelname (\DAE) is then used to filter each predicted pose by a 3-layer LSTM network, reducing accumulation of correlated error and hence drift over time. Furthermore to alleviate insufficiency of commonly used quality metric, we propose a new evaluation protocol using action classifiers to assess the quality of synthetic motion sequences. The proposed protocol can be used to assess quality of generated sequences of arbitrary length. Finally, we evaluate our proposed method on two of the largest motion-capture datasets available and show that our model outperforms the state-of-the-art techniques on a variety of actions, including cyclic and acyclic motion, and that it can produce natural looking sequences over longer time horizons than previous methods.
\end{abstract}

\section{Introduction}\label{sec:introdcution}
Predicting human motion over a significant time horizon is a challenging problem with applications in a variety of domains. For example in human computer interaction, human detection and tracking, activity recognition, robotics and image based pose estimation it is important to model and predict the most probable sequence of human motions in order to react accordingly and in a timely manner. Despite the inherent stochasticity and context dependency of natural motion, human observers are remarkably good at predicting what is going to happen next, exploiting assumptions about continuity and regularity in natural motion. However, formulating this domain knowledge into strong predictive models has been proven to be difficult. Integrating spatio-temporal information into algorithmic frameworks for motion prediction is hence either done via simple approximations such as optical flow~\cite{fragkiadaki2013pose,ramanan2005strike} or via manually designed and activity specific spatio-temporal graphs~\cite{ferrari2008progressive,SRNN_paper}. Given the learning capability of deep neural networks and recurrent architectures in particular, there lies enormous potential but also many challenges in learning statistical motion models directly from data that can generalize over a range of activities and over long time horizons.

Embracing this challenge we propose a new augmented recurrent neural network (RNN) architecture, dubbed \modelnamefull (\DAELSTMthreeLR). Our model is capable of extracting both structural and temporal dependencies directly from the training data and does not require expert designed and task dependent spatio-temporal graphs for input as is the case in prior work~\cite{SRNN_paper}. Our work treats the two aspects of the task, namely the inherent constraints imposed by the skeletal configuration and the constraints imposed by temporal coherence explicitly. Using a feed forward network for pose filtering and an RNN for temporal filtering, reduces drift due accumulation of error over time. We demonstrate this in a number of side-by-side comparisons to the state-of-the-art.

Specifically, we leverage de-noising autoencoders to learn the spatial structure and dependencies between different joints of the human skeleton while an LSTM network models temporal aspects of the motion. Contrary to related work that uses autoencoders to project the input data into a lower-dimensional manifold \cite{ERD_paper,SRNN_paper}, our model directly operates in the joint angle domain of the human skeleton. Although we use an autoencoder-like architecture it does not bear real resemblance to encoding-decoding in the usual sense of latent representation learning. We simply use the autoencoder to de-noise skeletal poses at every time step, i.e. our auto encoder takes a pose as input and produces the filtered version of it in the same domain. During training we perturb the inputs with random noise, as is common practice in de-noising tasks, but additionally use dropout layers on the inputs to randomly remove entire joint positions from the training samples. Therefore, to be able to accurately reconstruct entire poses the network has to leverage information about the spatial dependencies between adjacent joints to correctly infer positions of the missing joints. Hence this training regime forces the network to implicitly recover the spatial configuration of the skeleton. 

The proposed model learns to predict the most likely pose at time $t+1$ given the history of poses up to time $t$. Putting this model into recurrence allows for synthesis of novel and realistic motion sequences. We experimentally demonstrate that separating pose reconstruction and temporal modeling improves performance over settings where the autoencoder is primarily used for representation learning.  While the architecture is simple, it captures both the spatial and temporal components of the problem well and improves prediction accuracy compared to the state-of-the-art on two publicly available datasets.

In the domain of generative motion models, the lack of appropriate evaluation protocols to asses the quality and naturalness of the generated sequences is a commonly faced issue. The generated sequences need to be perceptually similar to the training data but clearly one does not simply want to memorize and replicate the training data. To better assess this generative nature of the task we furthermore contribute an evaluation protocol that quantifies how natural a generated sequence is over arbitrarily long time horizons. To assess naturalness we propose to train a separate classifier to predict action class labels. Intuitively the longer a sequence can be classified to belong to the same action category as the seed sequence the higher the quality of the prediction. 

We evaluate the proposed model on the H3.6m dataset of Ionescu et al. \cite{h36m_pami} and the more recent dataset of Holden et al. \cite{Holden:2016:DLF} in a pose forecasting task. Our model outperforms the 3-layer LSTM baseline and two state-of-the-art models \cite{ERD_paper,SRNN_paper} both in terms of short and long horizon predictions. Furthermore, we detail results from the proposed evaluation protocol and demonstrate that this can be used to analyze the performance of such generative tasks.

\begin{figure*}[t]
\begin{center}
   \includegraphics[clip, trim=0cm 9.8cm 0cm 0.5cm, width=0.9\textwidth]{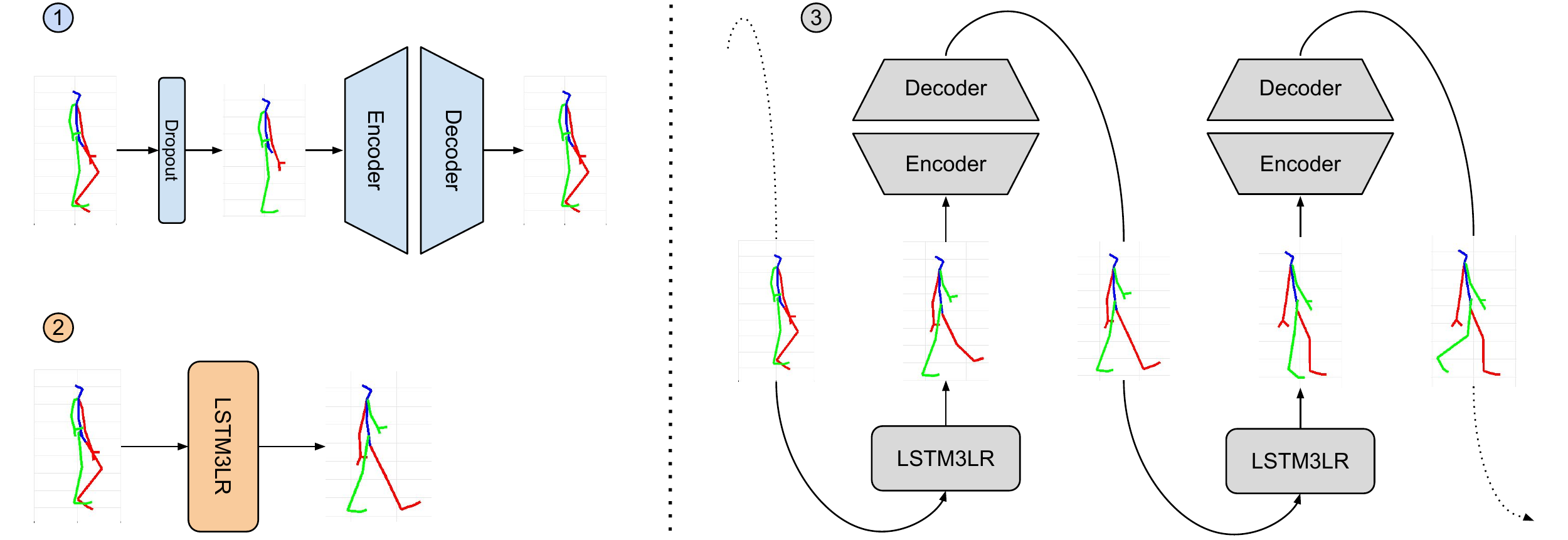}
\end{center}
   \caption{Schematic overview over the proposed method. (1) A variant of de-noising autoencoders learns the spatial configuration of the human skeleton via training with dropouts, removing entire joints at random which have to be reconstructed by the network. (2) We train a 3-layer LSTM recurrent neural network to predict skeletal configurations over time. (3) At inference time both components are stacked and the dropout autoencoder filters the noisy predictions of the LSTM layers, preventing accumulation of error and hence pose drift over time.}
\label{fig:DropoutEncoderLSTM}
\end{figure*}



\section{Related Work}
Here we provide an overview of recent literature that deals with human motion modeling. This is one of the core problems in computer vision and machine intelligence and has hence received much attention in the literature (for surveys see \cite{aggarwal1997human, mandery2016unifying, sagayam2017hand}. Recently deep learning based approaches have outperformed traditional methods on many body skeleton based tasks~\cite{han2016space} and hence we focus our discussion on motion prediction via deep learning methods. 

Spatio-temporal modeling of human activity is a crucial aspect in many problem domains including activity recognition from videos \cite{liu2016spatio}, human-object interaction \cite{Anticipating_Human_Activities} and robotics \cite{butepage2017anticipating}.
Manually designed spatio-temporal graphs (st-graphs) are typically applied to represent such problems, where nodes of the graph represent the interaction components, and the edges capture their spatio-temporal relationship. However, creating these models requires expertise and domain knowledge. Holden \etal \cite{Holden:2016:DLF} and Judith \etal \cite{DeepRepreSentationLEarning} propose a generative model for the automation of character animation in graphics. However, this approach is not predictive in the sense of prior poses and hence is not suitable for many vision tasks. 

In particular the activity and action recognition communities have explored the use of spatio-temporal models for image based action recognition \cite{du2015hierarchical,liu2016spatio,veeriah2015differential} and human object interaction \cite{Anticipating_Human_Activities, Human_Object_Interactions}. Often several different networks are trained separately and connected manually whereas we learn spatial structure and the spatio-temporal aspects in an end-to-end trainable model and directly from data. The task of motion prediction or motion synthesis is a relatively recent development and has seen comparatively little attention in the literature \cite{ERD_paper,SRNN_paper}.

Generally speaking there are two main directions in modeling temporal dependencies and state transitions. Namely, explicit parametric filtering such as Gaussian processes or other variants of Bayesian filtering such as HMMs or the Kalman Filter~\cite{Gaussian_Process_Dynamical_Models, wu2014leveraging}. Alternatively, various flavors of deep learning methods and in particular recurrent neural networks (RNNs) have been proposed for a variety of tasks~\cite{graves2013generating,hinton2012deep,LSTM,vinyals2015show}. These methods currently outperform traditional methods in many domains including that of motion prediction, with the two methods proposed in \cite{ERD_paper,SRNN_paper} being the most closely related to ours.

Fragkiadaki \etal \cite{ERD_paper} propose to jointly learn a representation of pose data and its time variance from images. An autoencoder is used for representation learning while the time variance is learned through an RNN which is sandwiched between the encoder and the decoder. The main focus of the work is to extract motion from video frames where representation learning step is crucial. However, for body pose based motion prediction the joint angle space of the human skeleton is already relatively low dimensional and the sequences are smooth. Hence, in cases where the input is already available in joint angle form, we argue that an additional representation learning step is not necessary. 
In consequence, our method employs a spatio-temporal component that directly operates in joint-angle space, whereas the work in \cite{ERD_paper} operates on the transformed latent space. Specifically, we separate concerns where the autoencoder is used as a spatial filter and the RNN as temporal predictor. Furthermore, we propose a different learning strategy and architecture to minimize correlation between the predictor and the filter.

\PG{Using over complete autoencoders to model kinematic dependencies imposed by human skeleton has been propsed for image based cases \cite{Structured_Prediction_of_3D_Human_Pose}. In contrast, our approach does not model the temporal dependency in the latent space of the autoencoder. This is motivated by the observation that unlike image data, mocap data in its original representation is smooth and continuous while there exists no guarantees of these properties in the learnt latent space.}

Martinez \etal \cite{MJB_on_human_motion} treat the problem of human motion modeling, focusing on short term action prediction and conclude that achieving both long and short term accuracy remains challenging. This is accredited to side-effects of curriculum learning, degrading short term prediction results. \cite{MJB_on_human_motion} avoids long term prediction and only reports results for a maximum of 400 ms into the future which is arguably sufficient for articulated object tracking but may not be sufficient for other tasks. In our work, decoupling spatial and temporal filtering during training improves robustness of the network over long time horizons, while maintaining short term prediction accuracy.

Integration of structural information in the form of spatio-temporal structural elements into deep learning models has been attempted before in \cite{SRNN_paper,liu2016spatio,NTU_RGBD}. This often requires manual design of structural elements. The main focus of Jain \etal \cite{SRNN_paper} is to automate the transformation of manually created st-graphs into an LSTM architecture. Although this process removes much manual labor it introduces a multitude of hyper parameters, such as network architecture and design for every independent node and edge. Further more due to inherent constraints in such networks they are usually less powerful than an unstructured network of similar size. This necessitates \cite{SRNN_paper} to train different models for different activities even within the H3.6M dataset. While our work also leverages the spatial structure of the data, we propose a method that does not require expert designed nor action specific st-graphs as input but instead learns the spatial structure of the skeleton directly from the data. The key idea is to train a deep autoencoder network to implicitly capture the inter-joint dependencies by randomly removing individual joints from the inputs at training time. The temporal evolution of the motion sequences is captured by an LSTM network operating directly on reconstructed and de-noised poses. Contrary to previous work \cite{Holden:2016:DLF, SRNN_paper} we train a single, unified model to perform all actions and do not require task specific networks.


\section{Method}
Figure \ref{fig:DropoutEncoderLSTM} illustrates our proposed architecture. The method comprises of two main components, namely, a \modelname (\DAE) and $3$-layer LSTM (\LSTMpart). These components serve distinct purposes but accomplish a common task, that of predicting human motion into the future. More precisely the model predicts a pose configuration for the time step $X_{t+1}$ given all $X_{1:t}$ prior poses up to time step $t$. Each pose at time $t$ consists of joint angles $X_t = [x_1, x_2, ... x_n]$ of the human skeleton. Hereby the \LSTMpartSpace outputs the most probable pose $X_{t+1}$ given $X_{1:t}$. The result is then fed to an autoencoder acting as a filter to refine the prediction based on the implicit constraints imposed by the human skeleton.

The main motivation and novelty in our approach is threefold. First, the data underlying our task has a well defined spatial structure in the human skeleton and integrating this domain knowledge into the model is important. We focus on data driven recovery of the spatial configuration unlike previous attempts \cite{SRNN_paper} which model it manually. Second, we observe that human motion is typically smooth, low dimensional and displays consistent spatio-temporal patterns. Hence we make no effort to perform representation learning which can potentially introduce detrimental artifacts. Third, at inference time the predicted poses recursively serve as input for the next time step and hence even small errors in the prediction will quickly accumulate and degrade the prediction quality over long time horizons. To avoid this we de-correlate errors in each time step with output from two networks with widely different characteristics.

With these observations in place we propose a simple yet effective network architecture comprising of two main components dedicated to learning and modeling the structural aspects of the task and the spatio-temporal patterns respectively. An autoencoder learns to model the configuration of the human skeleton and is used to filter noisy predictions of the RNN but only operates in the spatial domain. 

During training, both the autoencoder and 3-layer LSTM networks are pre-trained independently. In a subsequent fine-tuning step both models are trained further in an end-to-end fashion.

\subsection{Learning spatial joint angle configurations}
The \modelname (\DAE) component is based on de-noising autoencoders, used for the learning of representations that are robust to noisy data \cite{vincent2008extracting}. More formally, a de-noising autoencoder learns the conditional distribution $P_{\theta_D}(X|\tilde{X})$ where $P_{\theta_D}$ is represented by a neural network with parameters $\theta_D$, to recover the data sample $X$ given a corrupted sample $\tilde{X}$. During training, $X$ is perturbed by a stochastic corruption process $C$ where $\tilde{X} \sim C(\tilde{X}|X)$ \cite{ybengio2013generalized}.

Similarly to prior work we perturb our input data with random noise but importantly extend the architecture to more explicitly reason about the spatial configuration of the human skeleton. We introduce dropout layers directly after the input layer with the effect of randomly removing joints entirely from the skeleton rather than simply perturbing their position and angles. The only way to recover the full pose $X$ is then to reconstruct the missing joint angle information via inference from the adjacent joints. Importantly, during pre-training of the \DAEspace we do not use any temporal information but for consistencies sake keep the time subscript $t$ in this section. For a pair of clean and corrupted pose samples $(X_t, \tilde{X_t})$ we minimize the squared Euclidean loss:

\begin{equation}
\mathcal{L}(\cdot) = \norm{X_t - \tilde{X_t}}^2 = \sum_n^N (x_n - \tilde{x}_n)^2
\label{euclideanError-\DAE}
\end{equation}

During training of \DAEspace the corruption process $C$ is implicitly modeled in the network by the dropout layer just after the input layer. Introducing the dropout layer directly after the input layer forces the network to implicitly learn the spatial correlation of joints and our experiments suggest that this scheme produces better results than using the more standard multivariate Gaussian de-noising scheme only.

\subsection{Learning temporal structure}
Our goal is to recursively predict natural human poses into the future given a seed-sequence of motion capture data. This task shares similarities with other time-sequence data such as handwriting synthesis for which RNNs augmented with LSTM memory cells~\cite{LSTM} have been shown to work well~\cite{graves2013generating}. Similar to prior work \cite{ERD_paper,SRNN_paper} we leverage a $3$-layer LSTM network to model the temporal aspects of the task and to predict the poses forward over the time horizon. Each predicted pose $X_{t+1}$ is filtered by the \DAEspace component before feeding it back into the \LSTMpartSpace network, improving the prediction quality and reducing drift over time.

The \LSTMpartSpace network can either be utilized as a Mixture of Density Network (MDN) for probabilistic or as usual for deterministic predictions \cite{bishop1994mixture}. In the probabilistic case the output is modeled by a distribution family $P_{\theta_L}(X_{t+1}|X_{1:t})$ such as a Gaussian Mixture Model (GMM). The network is then used to parametrize the predictive distribution and trained by minimizing the negative log-likelihood. In the deterministic case the predictive distribution $P_{\theta_L}(X_{t+1}|X_{1:t})$ is implicitly modeled by the \LSTMpartSpace network with parameters $\theta_L$. The network is trained by minimizing the Euclidean loss between target and predicted pose configuration.
\begin{equation}
\mathcal{L}(\cdot) = \norm{X_{t+1} - \hat{X}_{t+1}}^2 = \sum_n^N (x_n - \hat{x}_n)^2,
\label{euclideanError-LSTM}
\end{equation}
where $X_{t+1}$ and $\hat{X}_{t+1}$ are the ground truth and predicted pose for time step $t+1$ respectively.

In the case of handwriting synthesis~\cite{graves2013generating} the inputs are low-dimensional and sampling from a GMM distribution has been shown to prevent collapse to the mean sample. For higher dimensional data such as human poses used in this work it is only practical to use very few mixture models which furthermore have to be restricted to diagonal covariances for each component. The deterministic and probabilistic prediction configurations did not show any significant differences in our qualitative and quantitative experiments. Prior work reports similar relative performance of deterministic and probabilistic prediction \cite{ERD_paper}. Concurring with \cite{ERD_paper} we conclude that the expressive power of a mixture model with few components for high dimensional tasks such as the human motion prediction is actually inadequate and hence all but one mixture model component collapses essentially making just unimodal prediction and we hence chose the deterministic parametrization. Our experiments show that our model can produce more realistic locomotion sequence over longer time horizons than the state-of-the-art (cf. \ref{subsection:action-classes}).

\subsection{Training and inference}
As outlined above it is fair to expect that the \LSTMpartSpace component will start to predict at least somewhat noisy poses after a sufficiently large number of time steps. We therefore assume that the corruption process $C$ is implicitly attached to the LSTM network. Consequentially we leverage the \DAEspace component to filter and improve the prediction by counteracting the corruption process. Our final architecture is then formalized as:
\begin{equation}
\text{\LSTMpart:  }\tilde{X_t} \sim P_{\theta_L}(X_{t}|X_{1:t-1})
\label{lstm_sample}
\end{equation}
\begin{equation}
\text{\DAE:  } X_t \sim P_{\theta_D}(X_t|\tilde{X_t})
\label{dae_sample}
\end{equation}
Because $\tilde{X_t} \sim C(\tilde{X_t}|X_t)$ and the LSTM are assumed to be coupled, the predictions drawn from the \LSTMpartSpace network (Eq. \ref{lstm_sample}) are also assumed to be corrupted. This assumption can be verified experimentally.

After the separate pre-training phase we stack the \LSTMpartSpace and \DAEspace components together and continue training with a brief fine-tuning phase (i.e., training for $\sim2$ epochs) using both losses from Eq. \ref{euclideanError-\DAE} \& \ref{euclideanError-LSTM}. We experimentally found that removing the dropouts during this fine-tuning process improves the performance. Inline with the literature \cite{rennie2014annealed} we experimentally confirmed that annealing the dropout rate for both the input and intermediate dropout layers to zero yields the best performance.
Finally, in a departure from prior work \cite{ERD_paper} the input and output representations of both the \DAEspace and the \LSTMpartSpace are in the original joint angle space rather than the latent space of the autoencoder.

At inference time (Figure \ref{fig:DropoutEncoderLSTM}, (3)) the \DAEspace component refines each of the \LSTMpart's pose predictions, leveraging the implicitly learned spatial structure of the skeleton. Our experiments show that this architecture leads to better sequence predictions across a variety of actions.


\section{Experiments}
We evaluate our proposed model extensively on two large publicly available datasets by Ionescu \etal \cite{h36m_pami} and Holden \etal \cite{Holden:2016:DLF}, These datasets contain a large number of subjects, activities and serve as good testbed for natural human motion in varied conditions.

\subsection{Datasets}
As we train one network that generalizes to all the action categories as opposed to our most closely related work \cite{ERD_paper,SRNN_paper} where a new model is trained for every activity, it is slightly unfair to compare the test errors directly. Yet to facilitate ease of comparison with the state-of-the art we evaluate our method on the H3.6M dataset following \cite{ERD_paper,SRNN_paper} and conduct additional experiments on the dataset accumulated by Holden \etal. Further more since with our implementation of SRNN following the protocol outlined in \cite{SRNN_paper} we did not manage to obtain competitive results in the Holden dataset \cite{Holden:2016:DLF}, we exclude this model from our experiments in the following sections. This could partially be because of lack of action labels in this dataset and hence we tried to train one SRNN model for all of the activities as opposed to action specific models.

\noindent\textbf{Human3.6M}
\cite{IonescuSminchisescu11,h36m_pami} is currently the largest single dataset of high quality 3D joint positions. It consists of 15 action categories, performed by seven different professional actors and contains cyclic motions such as walking and non-cyclic activities. The actors are recorded with a Vicon motion capture system, providing high quality
3D body joint locations in the global coordinate frame sampled at 50 frames per second (fps).  We follow \cite{SRNN_paper, ERD_paper} and treat subject 5 in a leave-one-subject-out evaluation setting. The dataset is down sampled by $2$ in time domain in order to obtain an effective fps rate of $25$

\noindent\textbf{Holden \etal} \cite{Holden:2016:DLF} accumulated a large motion dataset from many
freely available databases \cite{CMU_dataset, hdm05, mhad} and augmented these with their own data. The dataset contains around six million frames of high quality motion capture data for a single character sampled at 120 fps. While the dataset does not contain action labels it covers an even wider range of poses and hence serves well as complementary test set. We follow the training preprocessing settings reported in \cite{Holden:2016:DLF} and reserve $20\%$ of the dataset for testing. Similar to preprocessing of H3.6M dataset we down sample this dataset by $4$ to get an effective fps of $30$

\subsection{Implementation Details}
\noindent\textbf{Data preprocessing}
The above datasets have been preprocessed \cite{Holden:2016:DLF, SRNN_paper}  to normalize skeleton size i.e. height difference across all actors. The H3.6M data is further preprocessed so that the relative joint angles are taken as input as detailed in \cite{NIPS2006_3078}. This ensures direct comparability with \cite{ERD_paper,SRNN_paper}. Finally, we normalize each feature into the range of $[0, 1]$ separately and scale inputs during prediction time with the shift and scale values computed from the training data.

\noindent\textbf{Training details}
The auto encoder uses $3$ dense layers with $3000$ units each. \PG{We do not enforce weight sharing between layers. We use Relu nonlinearity to encourage sparsity and use dropout and $l2$ regularization to prevent over-fitting.} The learning rates are initialized as $0.005$ for the first stage of training and dropped by a factor of $2$ every time when validation loss flattens out. In end-to-end training, a lower learning rate $0.0001$ is used. The dropout rate is set to $0.5$ for the first stage and slowly annealed when validation error stops decreasing. The \DAEspace and \LSTMpartSpace networks are initially trained for $20$ epochs. The unified end-to-end model typically starts to converge after two epochs of fine-tuning. \PG{As is common we also make use of curriculum learning to train both the autoencoder and the LSTM network. A Gaussian noise with variance schedule of $[0.01, 0.05, 0.1]$ was used while a dropout schedule of $[0.01, 0.02, 0.04, 0.08, 0.1]$ was used. During the fine turning phase a reverse schedule is used. The scheduling hyper parameters did not impact the final model quality significantly.} 

\subsection{Impact of the Dropout Autoencoder}
As proposed in the method section we provide direct evidence here that the dropout learning scheme makes predictions more resilient against noise introduced by the RNN over time. 

In Figure \ref{fig:increasingDropout}-a we compare pose reconstruction performance under different amounts of input corruption for there different autoencoder settings: our proposed model \DAEspace (DropoutNoise), a standard de-noising autoencoder GAE (GaussianNoise) and a vanilla autoencoder (Vanilla). Our \modelname configuration is more robust to increasing amount of corruption and recovers the noisy input with lower error rates.
\begin{figure}[t]
\begin{center}
   \includegraphics[width=1\linewidth]{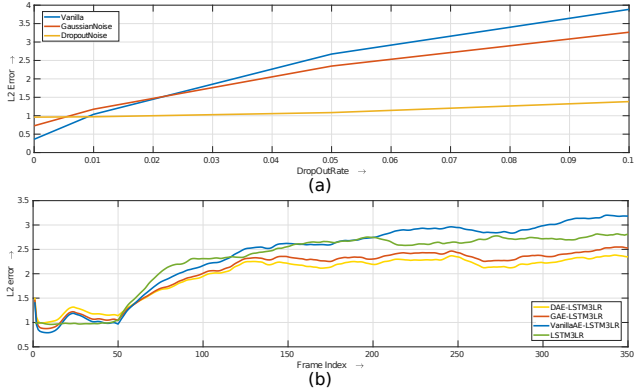}
\end{center}
   \caption{Pose reconstruction error (L2) over dropout rate (subplot above) and Effect of RNN output filtering (L2 Error vs timestep subplot below). Both standard curriculum learning and ours improve reconstruction but ours is more robust to large perturbations.}
   \label{fig:increasingDropout}
\end{figure}

Similarly, we compare the performance of these autoencoders by stacking them with a pre-trained \LSTMpartSpace network. The autoencoders are expected to filter out noisy predictions of the LSTM component. The filtered predictions are then compared with the ground-truth data. Figure \ref{fig:increasingDropout}-b shows that our model \DAELSTMthreeLR~ yields better performance and \DAEspace improves the prediction quality by effectively removing the noise introduced by the \LSTMpartSpace network.

Further, we asses the impact of the \DAEspace component on overall prediction accuracy in our second evaluation dataset. Table \ref{denoising_helps} compares Euclidean distance to ground-truth averaged across the Holden dataset. Note that both rows result from the same model, however the top-row is the error of the unfiltered LSTM output and the bottom row is the average error after filtering these predictions with the \DAEspace component. The \LSTMpartSpace produces noisy predictions which are improved by the \DAEspace (note that these accuracies are identical to bottom row in Table \ref{holden_euclidean}).

\begin{table}
\begin{center}
\centering
\begin{tabular}{|l|c|c|c|c|c|c}
\hline
Methods &  \multicolumn{2}{l|}{Short-term (ms)} & \multicolumn{3}{l|}{Long-term (ms)}\\
 & 80 & 160 & 320 & 560 & 1000\\
\hline
Ours without & 2.72 & 3.39 & 4.44 & 3.96 & 4.02 \\
Filtering & & & & & \\
\hline
\textbf{Ours} & 2.42 & 3.14 & 4.37 & 4.09  & 4.03\\
\hline
\end{tabular}
\end{center}
\caption{Comparison of average error in joint position at different time horizons on Holden. Error in $cm$ for unfiltered LSTM predictions (top) and that obtained via filtering with the \DAEspace network (bottom). Filtering via the \DAEspace network at every time-step improves accuracy and reduces long-term drift.}
\label{denoising_helps}
\end{table}

\subsection{Short-term motion prediction}\label{subsection:Motion prediction}
We first report quantitative results on H3.6M dataset using the same experiment configuration with \cite{ERD_paper,SRNN_paper}. The metric is taken from \cite{ERD_paper}, simply calculating the Euclidean distance between the predicted MOCAP vector and the ground truth. Please note that while this metric is useful to evaluate the short-term\footnote{We have indicated in the comparison tables what can be considered as short term.} predictions, it may yield deceiving assessments in longer time horizons since good models generate novel sequences, where deviations from ground truth are desired indeed. 
As reported previously \cite{ERD_paper,SRNN_paper}, it is worthwhile noting that the metric does not always correspond directly with the observed motion quality. The direct Euclidean error computation between the predicted and the ground truth MOCAP vector makes this metric less intuitive. In other words, a minor error in base frame (e.g., hip joint) angle can cause a large visual error, while the same error at a child node (e.g., wrist or ankle joints) would cause an insignificant effect. We therefore report results only on the action classes that have been reported on in the literature.

Furthermore since we were not able to replicate the performance of SRNN \cite{SRNN_paper} with one model for all actions\footnote{The original work on SRNN \cite{SRNN_paper} implements different model for every action} we avoid reporting suboptimal results and only compare with results previously presented in \cite{SRNN_paper}.

Analogously to the literature we also include a comparison to 3-layer LSTM architecture (\LSTMpart) as a baseline. In all our motion prediction experiments we initialize each model with $50$ seed frames and then start predicting $300$ frames (12s) into the future.

\begin{table}
\begin{center}
\centering
\begin{tabular}{|l|c|c|c|c|c|c}
\hline
Methods &  \multicolumn{2}{l|}{Short-term (ms)} & \multicolumn{3}{l|}{Long-term (ms)}\\
 & 80 & 160 & 320 & 560 & 1000\\
 \hline
\multicolumn{6}{|c|}{Walking activity}\\
\hline
\LSTMpart\cite{SRNN_paper} & 1.18 & 1.50 & 1.67 & 1.81 & 2.20\\
ERD \cite{SRNN_paper} & 1.30 & 1.56 & 1.84 & 2.00 & 2.38 \\
S-RNN \cite{SRNN_paper}  &  1.08&1.34&1.60&1.90&2.13\\
\textbf{Ours} & \textbf{1.00}&\textbf{1.11}&\textbf{1.39}&\textbf{1.55}&\textbf{1.39}\\
\hline
    \multicolumn{6}{|c|}{Eating activity}\\
\hline
    \LSTMpart\cite{SRNN_paper} & 1.36 & 1.79 & 2.29 & 2.49 & 2.82\\
    ERD \cite{SRNN_paper} & 1.66 & 1.93 & 2.28 & 2.36 & 2.41 \\
    S-RNN \cite{SRNN_paper}  &  1.35&1.71&2.12&2.28&2.58\\
    \textbf{Ours} & \textbf{1.31}&\textbf{1.49}&\textbf{1.86}&\textbf{1.76}&\textbf{2.01}\\
    \hline
    \multicolumn{6}{|c|}{Smoking activity}\\
\hline
    \LSTMpart\cite{SRNN_paper} & 2.05 & 2.34 & 3.10 & 3.24 & 3.42\\
    ERD \cite{SRNN_paper} & 2.34 & 2.74 & 3.73 & 3.68 & 3.82 \\
    S-RNN \cite{SRNN_paper}  &  1.90&2.30&2.90&3.21&3.23\\
    \textbf{Ours} & \textbf{0.92}&\textbf{1.03}&\textbf{1.15}&\textbf{1.38}&\textbf{1.77}\\
    \hline
    \multicolumn{6}{|c|}{Discussion activity}\\
\hline
    \LSTMpart\cite{SRNN_paper} & 2.25 & 2.33 & 2.45 & 2.48 & 2.93\\
    ERD \cite{SRNN_paper}  & 2.67 & 2.97 & 3.23 & 3.47 & 2.92 \\
    S-RNN \cite{SRNN_paper}  &  1.67  &  2.03   & 2.20  & 2.39 & 2.43\\
    \textbf{Ours} & \textbf{1.11} & \textbf{1.20} & \textbf{1.38} & \textbf{1.53} & \textbf{1.73}\\
\hline
\end{tabular}
\end{center}
\caption{Comparison of short-term predictions ($<$1s) of the different models over four different activities on the H3.6M dataset. We report error as the Euclidean norm (L2) of un-normalized ground truth and predicted MOCAP vectors.}
\label{h36M_euclidean}
\end{table}
Table \ref{h36M_euclidean} summarizes results from the walking, eating, smoking and discussion activities for short- and long-term periods. The simple baseline (\LSTMpart) is surprisingly competitive in making short-term predictions. However, a qualitative inspection on Figure \ref{QualitativeComp} shows that the baseline quickly converges to the safe mean pose, whereas the other models generate diverse and natural poses. Since ERD does not explicitly model the skeletal structure it starts to generate unnatural poses quickly. Our model, on the other hand, continues to predict smooth and natural looking poses especially over the longest horizon (1000ms).

\begin{table}
\begin{center}
\centering
\begin{tabular}{|l|c|c|c|c|c|c}
\hline
Methods &  \multicolumn{2}{l|}{Short-term (ms)} & \multicolumn{3}{l|}{Long-term (ms)}\\
 & 80 & 160 & 320 & 560 & 1000\\
\hline
\LSTMpart & 2.76 & 3.41 & \textbf{4.23} & \textbf{3.89} & 4.12\\
ERD & 2.87 & 3.88 & 5.64 & 6.08 & 6.96 \\
\textbf{Ours} & \textbf{2.42}&\textbf{3.14}&4.37 & 4.09  & \textbf{4.03}\\
\hline
\end{tabular}
\end{center}
\caption{Comparison of short-term predictions ($<$1s) of the different models on the Holden dataset. We report average Euclidean norm (L2) error of ground truth and predicted MOCAP vectors (expressed in cm/joint unit). The height of the skeleton (1.7m) was used to convert errors to metric scale.}
\label{holden_euclidean}
\end{table}

\begin{figure}[t]
\begin{center}
   \includegraphics[width=1.2\linewidth, trim={4.8cm 6.8cm 2cm 6.8cm}, clip]{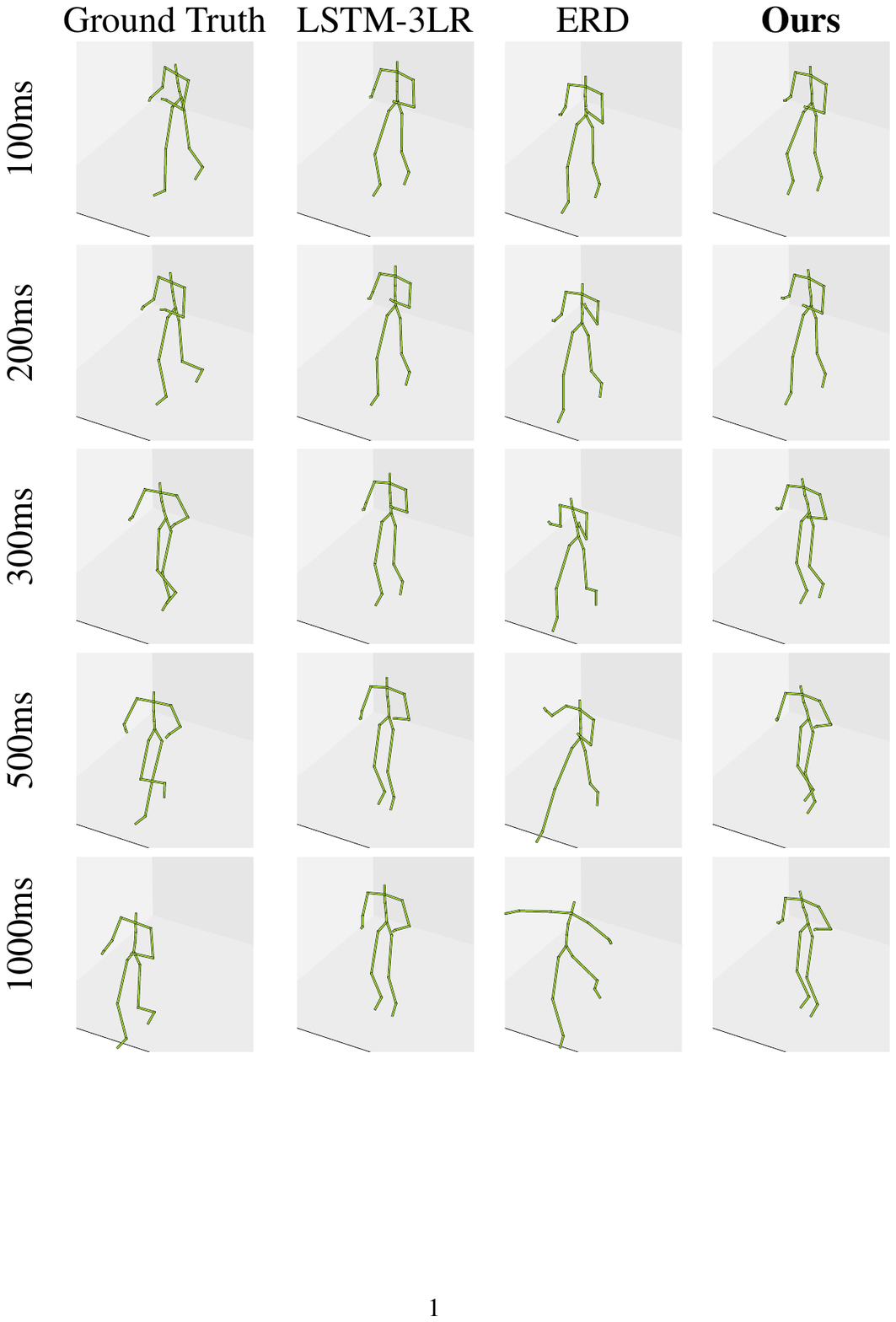}
\end{center}
   \caption{Qualitative comparison of our model with the state-of-the-art on the ``walking" activity. The baseline LSTM quickly converges to a safe mean configuration. ERD produces unnatural poses for longer horizons. Ours produces natural looking configurations without collapsing to an average pose.}
\label{QualitativeComp}
\end{figure}

Table \ref{holden_euclidean} shows the results obtained from the same experiment conducted on the Holden dataset. Because there are no action labels we average the error across all test sequences. Note that here the error has a unit of \textit{cm-per-joint} as opposed to \textit{unit distance} in the exponential domain in Table \ref{h36M_euclidean}. Similarly to the H3.6M case our model either outperforms others or performs similarly with the baseline model.

\subsection{A metric for motion synthesis}\label{subsection:action-classes}
In order to better differentiate model capabilities, especially for long-term prediction horizons, we leverage a pre-trained activity classifier for the evaluation of synthetic motion sequences. Intuitively, a high quality synthetic sequence should be assigned the same action label as the seed, whereas drift and motion degradation should impact the classification outcome negatively. This evaluation protocol is similar to the evaluation of generative adversarial networks \cite{odena2016conditional}. 
The evaluation by a separate classifier network is highly correlated to human judgment of action quality. Please refer to the supplementary videos that visualize action class probabilities of the classifier alongside an animated skeleton.

Given the success of the classifier in evaluating the synthetic sequences accurately, reformulating the problem as a auto-regressive generative adversarial network (GAN)  holds potential. However, it arguably requires significant modifications and we leave this as an interesting direction for future work. \PG{Here we provide a visual representation of the action classification probabilities and not their numerical values since their precise values are dependent upon the training details of the classifier, consequently making the precise probability values unimportant or even misleading.}

\begin{figure}[t]
\begin{center}
   \includegraphics[width=1\linewidth]{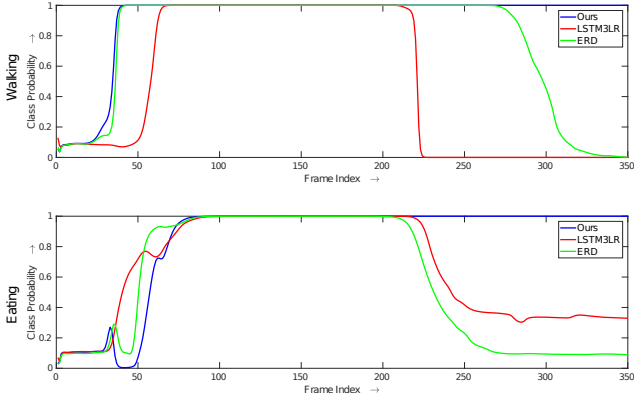}
\end{center}
   \caption{Comparison of class probabilities of long term sequence generation using pre-trained classifier. Our model generates sequences which belong to the same (correct) class for long $>10s$ horizons. The eating activity is initially confused with highly similar sitting activity but ours still yields best results.)}
\label{fig:ClassProbabilityOfNets}
\end{figure}

In our experiments we train a separate LSTM network performing on par with state-of-the-art action recognition methods \cite{NTU_RGBD}. It is used to assign class probabilities to the synthetic pose sequences generated by the baseline, the ERD network and our model.

\begin{figure}[t]
\begin{center}
   \includegraphics[width=1\linewidth]{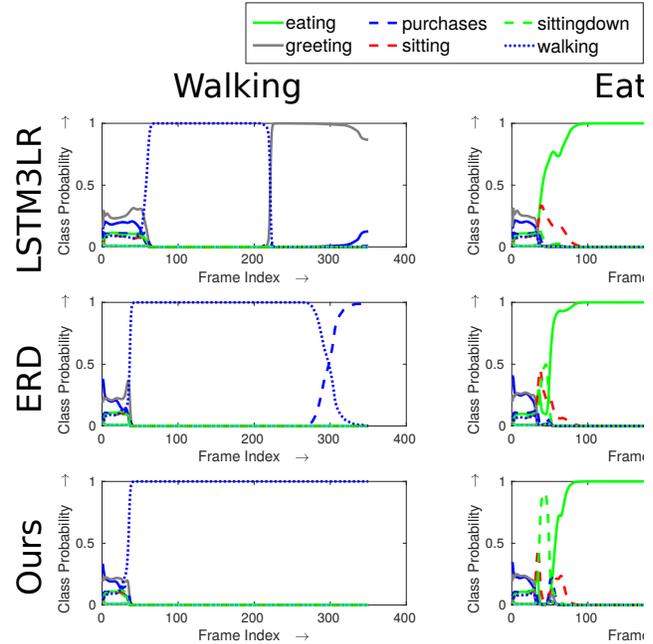}
\end{center}
   \caption{Multi-class probabilities for eating and walking. Initially in eating activity generated by our model is confused with sitting but eventually the arm motions lead to correct classification, whereas the baseline converges to a mean pose and hence remains ambiguous.}
\label{fig:ActionClasses}
\end{figure}

Figure \ref{fig:ClassProbabilityOfNets} plots class probabilities of ``walking'' and ``eating'' categories. Our model produces sequences that are classified correctly for longer time horizons than the baseline and ERD networks especially for cyclic motions such as ``walking''. Note that for the non-cyclic ``eating'' motion (Figure \ref{fig:ClassProbabilityOfNets}, bottom) the performance is degraded. Inspecting Figure \ref{fig:ActionClasses} closely reveals that the output from our model is initially confused with a very similar ``sitting" activity which only become distinguishable from ``eating'' when the hands start moving. This effect is best viewed in the video provided along with the supplementary material\footnote{As opposed to previous attempts we do not include this analysis for discussion and smoking action as they are indistinguishable from other activities in the action set (e.g. directions, waiting, walking and sitting) which confuses the classifier.}.

\section{Conclusion}\label{sec:conclusion}
In this paper we have proposed the \modelnamefull (\DAELSTMthreeLR) model for prediction of natural and realistic human motion given a short seed sequence. Our proposed model consists of a 3-layer LSTM (\LSTMpart) and a dropout autoencoder (\DAE). We train the autoencoder by randomly removing individual joints from the training poses in order to learn the spatial dependencies of the human skeleton implicitly. Furthermore, we have introduced an evaluation protocol that can be used to better analyze the quality of synthetic motion sequences in particular over long-time horizons where the simple Euclidean distance to the seed sequence does not provide a meaningful assessment anymore. Finally, we have experimentally demonstrated that our method outperforms the \LSTMpart baseline as well as the most closely related work ERD in a variety of experiments performed on two datasets. \PG{However if scrutinized closely one can notice that the animated skeleton leans slightly backwards. We attribute this to the fact that no physics based feedback is given to the model. Hence the model has no concept of mass, balance or gravity, which prevents it form identifying small error in overall orientation which strikes to human evaluator as physically impossible or improbable pose. Incorporating a physical model is left for future work.}


{\small
\bibliographystyle{ieee}
\bibliography{egbib}
}
\clearpage
\title{Supplementary for Learning Human Motion Models for Long-term Predictions}
\maketitle
\section*{Supplementary}
This document contains supplementary contributions complementing our main submission. Here we provide training details and additional experiments evaluating the efficacy of the proposed model in long-term motion sequence prediction. In particular, we detail the impact of the training scheme and the benefit of filtering the noisy LSTM output at every time step. We refer to the video for a qualitative comparison of motion predictions in longer horizons.

\section{Dropout autoencoder Training}
 We proceed training in a two stage process. First we train \LSTMpartSpace and the Dropout encoder separately. Training accuracy at this stage is not very important as this will follow a fine tuning process and premature stopping at this stage simply would result in a longer training time during the fine tuning stage. In all stages of training it is stopped once validation error converges. The noise $\sigma$ schedule for Curriculum Learning and Dropout Curriculum Learning was $[0.01, 0.05, 0.1]$ while the dropout schedule was $[0.01, 0.02, 0.04, 0.08, 0.1]$. For each configuration equal number of epochs is allocated from the budgeted epochs (usually ~10 - 15 epochs). Learning rate is set to $0.005$ initially and it is decreased by a factor of $2$ when the validation error plateaus. During the fine tuning stage we gradually decrease the noise level by following a reverse schedule e.g. ($[0.1, 0.08, 0.04, 0.02, 0.01, 0]$). These hyper-parameters are decided after conducting various experiments. 
 
Figure \ref{fig:RnadomNoiseCorruptionall} demonstrates that our auto encoder can recover to plausible poses from drastically distorted initial pose. Note that the recovered poses are not identical to the original ones, yet they look natural. Our \modelnamenbsp is capable of recovering the noisy poses naturally, which prevents \LSTMpartSpace from accumulating errors drastically.  

 \begin{figure}[H]
\begin{center}
   \includegraphics[width=1\linewidth]{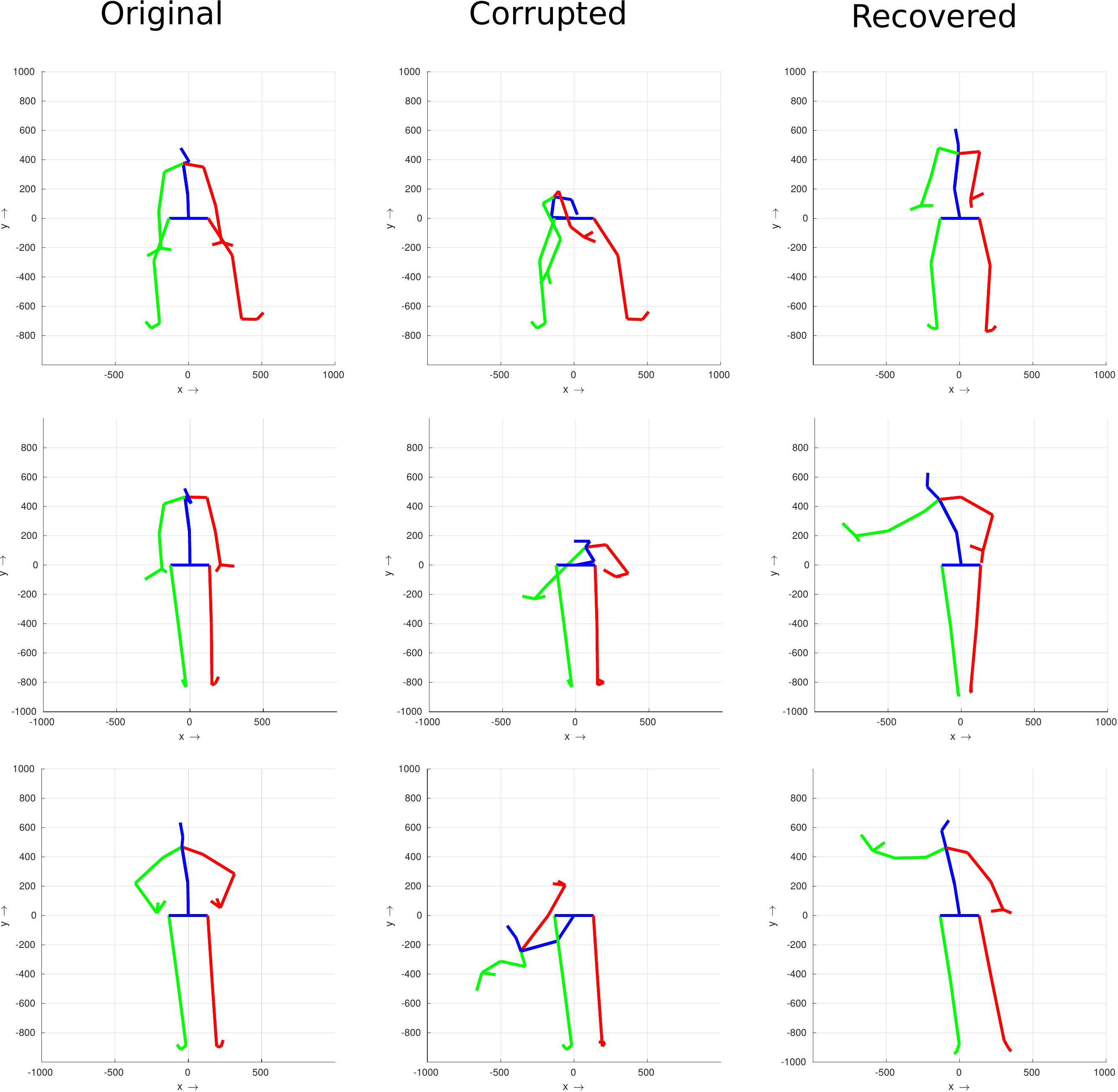}
\end{center}
   \caption{Original, corrupted and recovered human skeleton poses. Our \modelnamenbsp recovers poses naturally despite the fact that the recovered poses are not identical with the original poses.}
   \label{fig:RnadomNoiseCorruptionall}
\end{figure}
 
\section{Long-term motion prediction}
In the supplementary video it can be seen that \LSTMpartSpace converges to a mean pose and ERD drifts to unnatural poses, while our model continues generating natural walking sequence (from 00:10 to 00:35). Moreover, our model combines walking and drinking activities naturally. For aperiodic tasks such as eating our model keeps generating plausible poses (from 01:10 to 01:32). Please note that since yaw angle is represented as velocity in the data, the mean poses that models converge tend to rotate in yaw because of the integrated error. This is particularly visible in \LSTMpartSpace and ERD's predictions. The similarity among the short-term predictions show that models are able to extrapolate the seed sequence into future naturally. Please find our code in \href{https://bitbucket.org/parthaEth/humanposeprediction/overview}{Bitbucket}\footnote{https://bitbucket.org/parthaEth/humanposeprediction/overview}.

\section{Action class probabilities}
Due to stochasticity in human motion direct comparisons between the predicted and ground-truth motions can be misleading. The quantitative comparisons may not reflect quality of the predictions. Instead, the high-level properties such as fluidity and naturality should be evaluated in order to judge the quality of a model. Hence, we prefer using a separate action classifier in our evaluations instead of providing euclidean error on ground truth samples.

As discussed in the paper, the supplementary video plots class probabilities alongside the animated skeleton sequences (from 00:47 to 01:30). In the beginning, the classifier gathers state and hence distributes similar probability mass to every class. As soon as the distinctive features are visible, it assigns the corresponding class with very high confidence.

\section{Extensions}
Controlling orientation of the pose by means of external inputs is left as an interesting future work. As it can be seen in the video, our model is able to follow the user inputs despite the fact that it hasn't trained for this task (from 01:43 to 02:38). We show that a humanoid skeleton can be driven in any directions by user provided global orientation. This indicates that the proposed method can be useful in different types of use cases including motion prediction and real-time synthesis for character animation.

\end{document}
\fi